\pdfoutput=1

\documentclass[11pt]{article}

\usepackage{ACL2023}

\usepackage{times}
\usepackage{latexsym}

\usepackage{cleveref}

\usepackage[T1]{fontenc}

\usepackage[utf8]{inputenc}

\usepackage{microtype}

\usepackage{inconsolata}

\title{ThangDLU at \#SMM4H 2024: Encoder-decoder models for classifying text data on social disorders in children and adolescents}

\author{Hoang-Thang Ta \\
  Dalat University \\
  \texttt{thangth@dlu.edu.vn} \\\And
  Abu Bakar Siddiqur Rahman \\
  University of Nebraska at Omaha \\
  \texttt{abubakarsiddiqurra@unomaha.edu } \AND
  Lotfollah Najjar \\
  University of Nebraska at Omaha \\
  \texttt{lnajjar@unomaha.edu} \\\And
  Alexander Gelbukh \\
  Instituto Politécnico Nacional (IPN), Mexico \\
  \texttt{gelbukh@cic.ipn.mx}
  }

\begin{document}
\maketitle
\begin{abstract}
This paper describes our participation in Task 3 and Task 5 of the \#SMM4H (Social Media Mining for Health) 2024 Workshop, explicitly targeting the classification challenges within tweet data. Task 3 is a multi-class classification task centered on tweets discussing the impact of outdoor environments on symptoms of social anxiety. Task 5 involves a binary classification task focusing on tweets reporting medical disorders in children. We applied transfer learning from pre-trained encoder-decoder models such as BART-base and T5-small to identify the labels of a set of given tweets. We also presented some data augmentation methods to see their impact on the model performance. Finally, the systems obtained the best F1 score of 0.627 in Task 3 and the best F1 score of 0.841 in Task 5.
\end{abstract}

\section{Introduction}
Social disorders are significantly influencing a large proportion of young people globally. Social anxiety disorder (SAD) typically emerges during early adolescence and is characterized by excessive anxiety in social situations~\cite{rao2007social}. Although spending time outdoors in green or blue environments has been shown to alleviate symptoms of various anxiety disorders, limited research has explored its impact specifically on SAD. Meanwhile, numerous children receive diagnoses of conditions that can significantly affect their daily functioning and persist into adulthood. The commonly diagnosed childhood disorders are attention-deficit/hyperactivity disorder (ADHD)~\cite{kidd2000attention}, autism spectrum disorders (ASD)~\cite{matson2009parent}, speech delay, and asthma. 

Datasets related to these disorders are usually extracted from tweets or user posts on social platforms such as Twitter and Reddit. Users describe their disorders daily and receive feedback from the community or comment on what they have passed. As a text classification problem, the most advanced and popular methods of social disorder identification use deep learning networks such as BERT~\cite{devlin2018bert}, RoBERTa~\cite{liu2019roberta}, Transformer models~\cite{vaswani2017attention}, and their variants. 

In participating in the \#SMM4H 2024 workshop~\cite{smm4h-2024-overview}, we apply transfer learning from two pre-trained models (BART-base~\cite{lewis2019bart} and T5-small~\cite{raffel2020exploring}), which follow the architecture of Transformer and sequence-to-sequence. Additionally, we exploited two data augmentation methods — (1) false inferred data and (2) paraphrased data extracted from ChatGPT — to supplement the training set and see their impact on model performance. Because the organizers hide the final ranking table, we can only present our results compared to the mean and median values by metrics (F1, precision, recall, and accuracy) they provided.

\section{Methodology}
After conducting several initial experiments on training the dataset with and without preprocessing steps, we observed that training without any preprocessing yielded better performance. It is assumed that every token within the data can positively impact the model's performance. Therefore, we fed raw data directly into the model in the training process.

We applied transfer learning from two pre-trained models based on the architecture of the Transformer and sequence-to-sequence: BART-base~\cite{vaswani2017attention} and T5-small~\cite{raffel2020exploring} for Task 3 and Task 5, respectively. They are encoder-decoder language models, producing outputs that can be unknown labels. From a given text, the models must detect its label, and any out-of-scope label will be set automatically to a default label, which we choose as "0" for both tasks. The best models were saved based on their performance on the validation set by F1-macro for Task 3 and F1-micro for Task 5.

Two data augmentation methods complement new data to improve mode performance. First, the false inferred data in the validation set were utilized when inferring them in a trained model with the default training data. Second, paraphrased data by ChatGPT is taken from the training and validation sets at an insignificant cost.

\section{Tasks \& Datasets}

\subsection{Task 3}
This task involves classifying Reddit posts mentioning predetermined keywords related to outdoor spaces into one of four categories: ("1") positive effect, ("2") neutral/no effect, ("3") negative effect, and ("4") unrelated \footnote{https://codalab.lisn.upsaclay.fr/competitions/18305}. The dataset comprises 3,000 annotated posts from the \texttt{r/socialanxiety} subreddit, filtered for users aged 12-25, and keywords related to green or blue spaces. 80\% of the data will be used for training/validation, and 20\% for evaluation. Evaluation will be based on the macro-averaged F1-score across all categories. Data will be provided in CSV format with fields: \texttt{post\_id}, \texttt{keyword}, \texttt{text}, and \texttt{label}. The distribution of subsets follows a ratio 6:2:2, in which training, validation, and testing sets take 1800, 600, and 600 posts correspondingly. However, the organizer provided a test set with 1200 posts to hide the real ones.

\subsection{Task 5}
This task involves automatically classifying tweets from users who reported pregnancy on Twitter. It distinguishes tweets reporting children with ADHD, autism, delayed speech, or asthma ("1") from those merely mentioning a disorder ("0")\footnote{https://codalab.lisn.upsaclay.fr/competitions/17310}. The goal is to enable large-scale epidemiologic studies and explore parents' experiences for targeted support interventions. The dataset includes 7398 training tweets, 389 validation tweets, and 1947 test tweets. Like Task 3, the organizer gave a new test with 10000 tweets to hide the actual data. 

\section{Experiments}
\subsection{Task 3}
\begin{table*}
\centering
\begin{tabular}{llllll}
\hline
\textbf{\#Submission}& \textbf{Data} & \textbf{F1} & \textbf{Precision}  & \textbf{Recall} &  \textbf{Accuracy} \\
\hline
1 & Training & 0.595 & 0.589 & 0.615 & 0.631 \\
2 & Training + Paraphrased & 0.601 & 0.592 & 0.622 & 0.640 \\
3 & \textbf{Training + Validation} & \textbf{0.627} & \textbf{0.620} & \textbf{0.644} & \textbf{0.670} \\
\hline
\multicolumn{5}{l}{\textit{Compared to other teams}} \\
- & Mean & 0.518 & 0.564 & 0.537 & 0.574 \\
- & Median & 0.579 & 0.630 & 0.588 & 0.627 \\
\hline
\end{tabular}
\caption{\label{tab_task_3}
F1-macro, Precison-macro, and Recall-macro values of BART-base models on Task 3, which were trained over different data combinations.
}
\end{table*}

The organizer limited each team to 3 submissions. Therefore, we used BART-base to train 3 models over 3 different training data sets. 

\begin{itemize}
    \item \texttt{Training}: The model was trained over the original training set offered by the organizer.
    \item \texttt{Training + Paraphrased}: We extracted the paraphrased data based on the validation set by ChatGPT. Then, we added this new data to the training set. 
    \item \texttt{Training + Validation}: We added a validation set to the training set and then used this new set for training the model. 
\end{itemize}

The training has the same parameters for all models, including \texttt{epochs = 10}, \texttt{batch\_size = 4}, and \texttt{max\_source\_length = 768}. For any input with a length over 768 tokens, we process it to take its first 256 tokens and its last 512 tokens. 

\Cref{tab_task_3} shows the results of our team and the mean and median performance of all teams by metrics: F1, precision, recall, and accuracy. It is clear our models outperformed the mean and median overall metrics. Our best model was trained on the \texttt{Training + Validation} data and obtained an F1 value of 0.627, while the model with \texttt{Training + Paraphrased} data takes slightly lower performance. While paraphrased data helps improve the model, it is better to collect actual data to obtain the best performance. The low F1 value indicates the task's difficulty and the need for adding more training data to improve the model performance.

\subsection{Task 5}
\begin{table*}
\centering
\begin{tabular}{lllll}
\hline
\textbf{\#Submission}& \textbf{Data} & \textbf{F1} & \textbf{Precision}  & \textbf{Recall}  \\
\hline
1 & Training + False inferred + Paraphrased  & 0.841 & 0.844 & 0.839 \\
2 & Training + False inferred & 0.829 & 0.803 & 0.856 \\
3* & Training + Validation & 0.870 & 0.869 & 0.867 \\
4* & Training & 0.820 & 0.809 &	0.831 \\
\hline
\multicolumn{5}{l}{\textit{Compared to other teams}} \\
- & Mean & 0.822 & 0.818 & 0.838 \\
- & Median & 0.901 & 0.885 & 0.917 \\
\hline
\multicolumn{5}{l}{\textit{Other works}} \\
- & \textbf{RoBERTa-Large}~\cite{klein2024using} & \textbf{0.930} & - & - \\
\hline
\multicolumn{5}{l}{*Our extra participation in the post-eval phase.} \\
\hline
\end{tabular}
\caption{\label{tab_task_5}
The metrics of T5-small models on Task 5, which were trained over different data combinations.
}
\end{table*}

The task limits each team to only 2 submissions. Therefore, we pick 2 trained T5-small models on two training sets for participation. In the post-eval phase, we also trained the other 2 models with different training sets. Finally, we have 4 models with 4 training sets, which are:

\begin{itemize}
    \item \texttt{Training}: The model was trained over the original training set offered by the organizer.
    \item \texttt{Training + Validation}: We added a validation set to the training set and then used this new set for training the model.
    \item \texttt{Training + False inferred}: First, we used the model trained on the original training set to infer the labels of inputs in the validation set. Then, we collect false inferred texts with their labels (41 examples) and add them to the original training set to form a new one.
    \item \texttt{Training + False inferred + Paraphrased}: Similar to \texttt{Training + False inferred}, we add more paraphrased data to the training set. First, we used BM25~\cite{robertson1995okapi} to take similar texts in the training set based on the validation set. Not that the new set's size equals the validation set's size (389 examples). Then, we used ChatGPT APIs\footnote{https://platform.openai.com/docs/overview} to extract paragraphed texts and add them to the training set.
    
\end{itemize}

The training has the same parameters for all models, including \texttt{epochs = 20}, \texttt{batch\_size = 4}, and \texttt{max\_source\_length = 128}. \Cref{tab_task_5} shows the results of our team and the mean and median performance of all teams by metrics: F1, precision, and recall. All our models have metric values that are better than the mean but lower than the median. Especially, our metric values are significantly lower than the benchmark F1 ~\cite{klein2024using} when using RoBERTa-Large. It can be explained that we only use a small-scale pre-trained model like T5-small for the classification task. 

Due to the small size of false inferred data, the model performance is not much better. However, we realize that the paraphrased data contributes positively to the model performance even though with a subset. Unfortunately, we can not experiment with training on more paraphrased data based on the full validation and the training sets, but we expect the model will be much better. Our best model was trained on the \texttt{Training + Validation} with an F1 value of 0.87, indicating that the more data, the better model performance.

\section{Conclusion}
This paper introduced our approach, utilizing pre-trained encoder-decoder models with two data augmentation methods to address Task 3 and Task 5 of the \#SMM4H 2024 workshop. Our findings underscore the advantages of encoder-decoder models in text classification problems when they offer a strong baseline performance. Furthermore, it is beneficial to exploit data augmentation methods to enhance the model's performance by complementing paraphrased texts from ChatGPT. According to the organizers, we officially achieved the highest F1 score of 0.627 for Task 3 and 0.841 for Task 5. In the future, we will investigate further how large language models' outputs like ChatGPT can positively impact downstream classification tasks' performance.

\newpage
\bibliography{anthology,custom}
\bibliographystyle{acl_natbib}




\end{document}